%% file: main.tex
\documentclass[journal=jacsat,manuscript=article]{achemso}

\usepackage[version=3]{mhchem} 
\usepackage{xcolor}
\usepackage{booktabs}
\usepackage{makecell}
\usepackage{graphicx}
\usepackage{subcaption}
\usepackage{float} 
\usepackage{threeparttable}
\usepackage{multirow}
\usepackage{hyperref}

\author{Filipp Nikitin}
\affiliation[CB]{Computational Biology Department, Carnegie Mellon University, Pittsburgh, PA, USA}
\alsoaffiliation[Chem]{Department of Chemistry, Carnegie Mellon University, Pittsburgh, PA, USA}
\altaffiliation{These authors contributed equally.}

\author{Ian Dunn}
\affiliation[UPitt]{Department of Computational and Systems Biology, University of Pittsburgh, Pittsburgh, PA, USA}
\altaffiliation{These authors contributed equally.}

\author{David Ryan Koes}
\affiliation[UPitt]{Department of Computational and Systems Biology, University of Pittsburgh, Pittsburgh, PA, USA}

\author{Olexandr Isayev}
\email{olexandr@olexandrisayev.com}
\alsoaffiliation[CB]{Computational Biology Department, Carnegie Mellon University, Pittsburgh, PA, USA}
\affiliation[Chem]{Department of Chemistry, Carnegie Mellon University, Pittsburgh, PA, USA}

\title[GEOM-Drugs Revisited]
  {GEOM-Drugs Revisited: Toward More Chemically Accurate Benchmarks for 3D Molecule Generation}

\abbreviations{IR,NMR,UV}

\begin{document}







\begin{abstract}
  Deep generative models have shown significant promise in generating valid 3D molecular structures, with the GEOM-Drugs dataset serving as a key benchmark. However, current evaluation protocols suffer from critical flaws, including incorrect valency definitions, bugs in bond order calculations, and reliance on force fields inconsistent with the reference data. In this work, we revisit GEOM-Drugs and propose a corrected evaluation framework: we identify and fix issues in data preprocessing, construct chemically accurate valency tables, and introduce a GFN2-xTB-based geometry and energy benchmark. We retrain and re-evaluate several leading models under this framework, providing updated performance metrics and practical recommendations for future benchmarking. Our results underscore the need for chemically rigorous evaluation practices in 3D molecular generation. Our recommended evaluation methods and GEOM-Drugs processing scripts are available at \url{https://github.com/isayevlab/geom-drugs-3dgen-evaluation}
\end{abstract}

\section{Introduction}

Generative models for molecules are an emerging paradigm that enables the construction of novel molecules in 2D or 3D~\cite{luo20213d,bilodeau2022generative}. These AI models learn the patterns and distribution of existing molecular data to generate previously unseen chemical structures. By encoding molecular information into mathematical representations and then sampling from a learned distribution, these models facilitate efficient exploration of vast chemical space. The field continues to evolve rapidly and is not yet mature. 

The field of cheminformatics has established fundamental protocols~\cite{cherkasov2014qsar,tropsha2010best} and best practices~\cite{bannwarth2019gfn2,artrith2021best} for achieving ML models with high statistical rigor and external predictive power~\cite{tropsha2010best}. Here, critical steps such as data preparation, chemical structure curation, outlier detection, dataset balancing, and rigorous ML model validation must be included into the overall data workflow. Multiple studies emphasized that chemical structure curation should be treated as a separate and critical component of any cheminformatics research~\cite{artrith2021best}. Seminal studies showed examples of how accumulation of errors and incorrect processing of chemical structures could lead to significant loss of accuracy of ML models~\cite{young2008chemical}. 

The GEOM data set~\cite{axelrod2022geom} is one of the most widely used large-scale high-accuracy datasets of molecular conformations. A subset of GEOM containing drug-like molecules, generally known as GEOM-Drugs, has become a foundational benchmark for developing 3D molecular generative models. The frequent use of GEOM-Drugs in this field has given rise to a somewhat standardized set of metrics to evaluate the quality of generative models trained on this dataset.
In this work, we identify several critical issues with the evaluation practices used in state-of-the-art 3D molecular generative models, which we believe are misleading the research community and limiting progress in the field.

First, we highlight three major problems with the commonly used ``molecular stability'' metric, which measures whether atoms have valid valencies. One of the original implementations contained a bug that causes chemically implausible valencies to be counted as valid, leading to inflated stability scores. This flawed implementation was reused by several follow-up works \cite{irwinsemlaflow,le2023navigating,dunn2024mixed,dunn2024exploring,reidenbach2024applications,vignac2023midi}, resulting in a significant body of work with misleading characterizations of model performance.


Second, many recent works lack rigorous and chemically grounded evaluation of 3D structures, which continues to hinder progress in generative modeling. Common issues include the use of oversimplified atom–atom distance lookup tables to evaluate the validity of generated 3D structures~\cite{huang2024learning, garcia2021n, hoogeboom2022equivariant, morehead2024geometry,song2023equivariant,xu2024geometric}, reliance on distribution-based metrics that are difficult to interpret~\cite{vignac2023midi,le2023navigating}, and the use of energy evaluations at inappropriate levels of theory, such as MMFF94, which is not suitable for assessing models trained on GFN2-xTB-optimized data~\cite{irwinsemlaflow,cornet2024equivariant}.

To address these issues, this paper provides:

\begin{enumerate}
\item A refined dataset split of GEOM-Drugs, which excludes molecules where GFN2-xTB calculations fractured the original molecule.
\item An updated molecule stability metric with a chemically accurate valency lookup table that is derived from this refined dataset.
\item An energy-based evaluation methodology for an accurate and chemically interpretable assessment of generated molecular 3D geometries.
\end{enumerate}

We retrained several widely used generative models on our reprocessed dataset and updated the evaluation metrics to address previously observed issues. Although the relative rankings of the models remained largely consistent, the updates yield practical improvements that highlight the critical importance of rigorous and accurate evaluation practices in the field.

\section{Molecule Stability}

Valency in chemistry refers to the combining capacity of an atom or element, describing how many chemical bonds it can form with other atoms. It is defined as the sum of bond orders of its covalent bonds. Due to chemical constraints (e.g., the octet rule), atoms of a given element and formal charge typically exhibit only a few plausible valencies; for instance, neutral carbon almost exclusively has a valency of 4. Molecules violating these valency constraints are chemically unstable. Thus, generative models must produce molecules adhering to these rules. A practical evaluation of generative models involves measuring the fraction of atoms with valid valencies, defined as valencies observed in the training data. A ``lookup table'' of valid valencies, consisting of tuples of (element, formal charge, valency), is created from the training set.

\begin{figure}[H]
  \centering
  \begin{subfigure}[t]{0.35\textwidth}
    \centering
    \includegraphics[width=\linewidth]{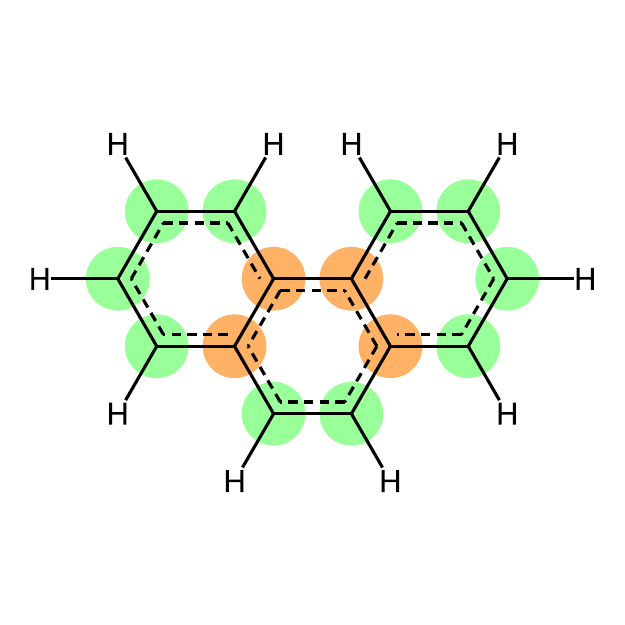}
    \caption{}
    \label{fig:tr_aromat}
  \end{subfigure}
  \hfill
  \begin{subfigure}[t]{0.35\textwidth}
    \centering
    \includegraphics[width=\linewidth]{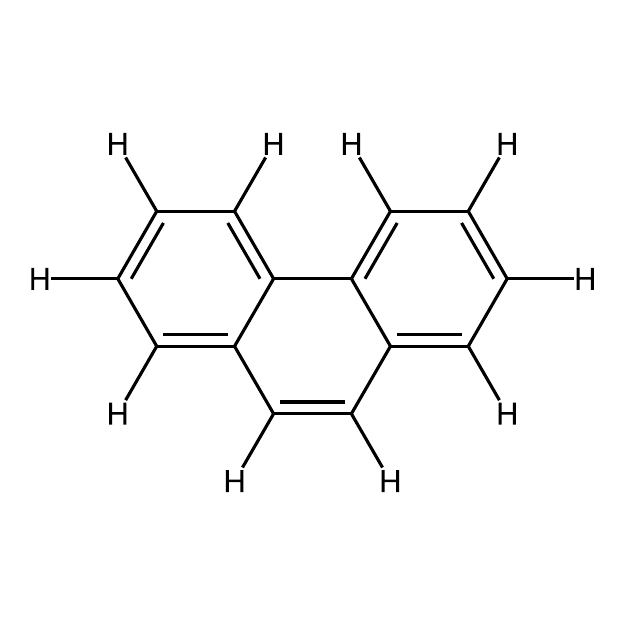}
    \caption{}
    \label{fig:tr_kekulized}
  \end{subfigure}
\caption{
An example of a molecule where the assumption that aromatic bonds contribute 1.5 to atomic valency holds only partially. In the aromatic form of triphenylene~(\subref{fig:tr_aromat}), the green-highlighted atoms are correctly classified as stable under the 1.5 assumption, while others are misclassified. In contrast, the kekulized representation~(\subref{fig:tr_kekulized}) resolves the ambiguity and yields chemically accurate valency assignments across all atoms. This illustrates the limitations of the 1.5 approximation in polycyclic aromatic systems.
}
  \label{fig:aromat_val}
\end{figure}

Valency can be computed as the sum of bond orders in a molecule’s Kekulized form, where bonds are explicitly represented as single, double, or triple. This approach works reliably for molecules without aromatic bonds. When aromatic bonds are introduced, however, valency computation becomes more complex. In simple cases such as benzene, one can assume each aromatic bond contributes 1.5 to the valency, yielding the correct total (e.g., carbon atoms in benzene are correctly assigned a valency of 4). But in more complex aromatic systems, this assumption may not hold, and valency contributions can vary depending on the bonding environment and resonance structures (see Figure~\ref{fig:aromat_val}).

 Initially, molecular stability was proposed in the EDM paper~\cite{hoogeboom2022equivariant}, where the authors argued for the evaluation of valency correctness directly on the raw output of generative models. They noted that traditional validity metrics, defined as the fraction of molecules that can be sanitized with RDKit, can be misleading, as RDKit may implicitly adjust hydrogen counts or modify aromaticity, altering the predicted molecule. We generally support the idea of assessing raw valencies, especially for models that explicitly generate both atoms and bonds because it provides a more chemically grounded evaluation. Unlike validity, stability captures whether the generated molecules respect elemental valence constraints without relying on post-processing.

\subsection{Identified Issues}


We identify multiple critical issues with the valency evaluation methods used in popular molecular generative models; these issues obscure instances where generative models produce chemically implausible structures. 

One of the pioneering models, MiDi, implemented a valency calculation method in which the valency contributions for all aromatic bonds were rounded to 1 instead of the intended value of 1.5. Thus, the valency computation for most atoms participating in aromatic bonds is incorrect. More importantly, it appears that the flawed valency computation was also used to construct the valency lookup table with which generated atoms are classified as ``stable'' or not, resulting in a lookup table with chemically implausible entries. For instance, the lookup table allows for neutral carbon with a valency of 3 and neutral nitrogen with a valency of 2. Implausible entries in the valency lookup table mask failures of the generative model and produce artificially inflated molecular stability values. Due to widespread reuse of MiDi’s code, this numerical error propagated to several works including EQGAT-Diff~\cite{le2023navigating}, SemlaFlow~\cite{irwinsemlaflow}, Megalodon~\cite{reidenbach2024applications}, and FlowMol~\cite{dunn2024exploring,dunn2024mixed}. Other models, such as JODO~\cite{huang2024learning} and NextMol~\cite{liu2025next}, computed valencies using an alternative approach based on RDKit kekulization. However, they still relied on an inappropriate lookup table for defining valid valency ranges. 

\begin{figure}[H]
  \centering
  \begin{subfigure}[t]{0.30\textwidth}
    \centering
    \includegraphics[width=\linewidth]{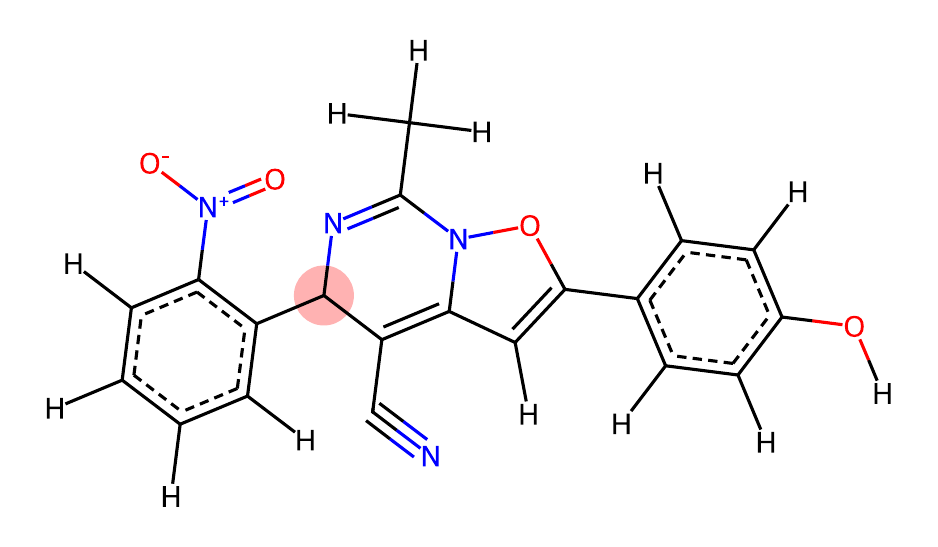}
    \caption{}
    \label{fig:molecule-a}
  \end{subfigure}
  \hfill
  \begin{subfigure}[t]{0.32\textwidth}
    \centering
    \includegraphics[width=\linewidth]{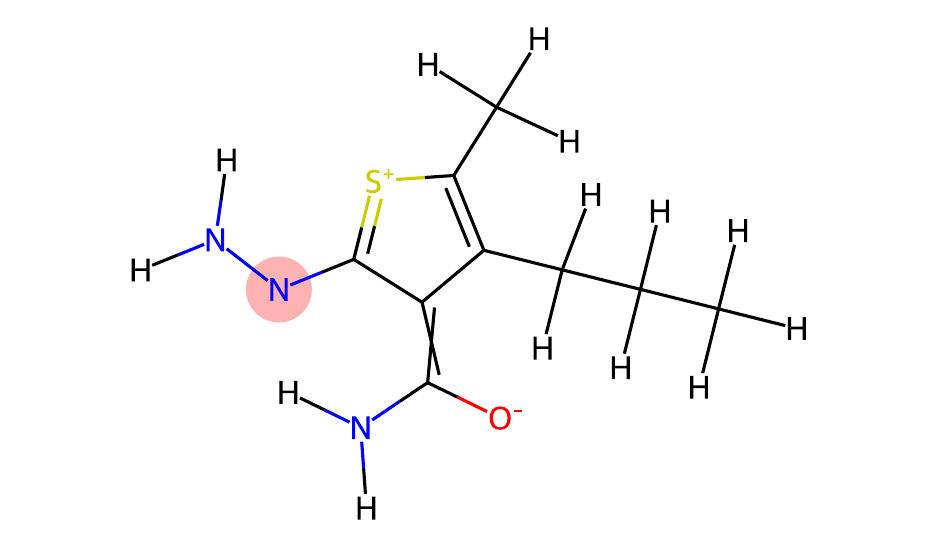}
    \caption{}
    \label{fig:molecule-b}
  \end{subfigure}
  \hfill
  \begin{subfigure}[t]{0.30\textwidth}
    \centering
    \includegraphics[width=\linewidth]{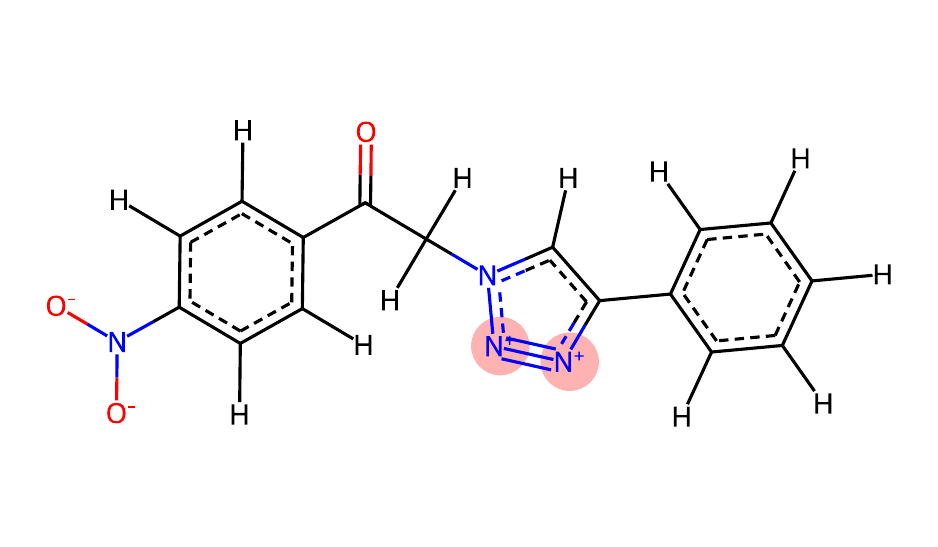}
    \caption{}
    \label{fig:molecule-c}
  \end{subfigure}

\caption{
Examples of molecules that pass the molecular stability test under commonly used evaluation criteria. These flawed metrics erroneously classify chemically invalid configurations as stable—including cases such as a neutral carbon with three single bonds~(\subref{fig:molecule-a}), a neutral nitrogen with two single bonds~(\subref{fig:molecule-b}), and a nitrogen atom with +1 charge bonded via both a triple bond and an aromatic bond~(\subref{fig:molecule-c}).
}
  \label{fig:stable-molecules}
\end{figure}

\subsection{Solution}

Two key solutions are necessary to correct the aforementioned problems with the molecular stability metric: fixing the valency computation bug for aromatic bonds and recomputing the valency lookup table. We quantify the effects of our proposed solutions by re-evaluating models that used the faulty molecular stability metric in their original publications: EQGAT-Diff \cite{le2023navigating}, Megalodon-quick \cite{reidenbach2024applications}, SemlaFlow \cite{irwinsemlaflow}, FlowMol2 \cite{dunn2024exploring}, and Megalodon-flow \cite{reidenbach2024applications}. The results of these reevaluations are shown in Table \ref{tbl:mol_stab}. All metrics were computed using 5,000 generated molecules per model. 

Correcting the numerical bug that erroneously rounded the contribution of aromatic bonds from 1.5 to 1 (without adjusting the lookup table) causes a dramatic drop in molecular stability. This can be observed by comparing the first two columns of Table \ref{tbl:mol_stab}. Additionally, this demonstrates that neither 1 nor 1.5 provides a universally reliable estimate for the contribution of an aromatic bond to atomic valency.


We propose two strategies to address the limitations in molecular stability computation. The first strategy involves enhancing the valency lookup table by explicitly accounting for aromaticity. Instead of the conventional tuples (element, formal charge, valency), we construct a more nuanced table indexed by (element, number of aromatic bonds, formal charge, valency), with the associated values representing allowed non-aromatic bond valencies—i.e., total bond order excluding contributions from aromatic bonds (see SI Table~\ref{tab:geom_drugs_tuple_valencies_hier}). In this formulation, each atom’s bonding environment is described by the tuple $(n_{\text{arom}}, v_{\text{other}})$, where $n_{\text{arom}}$ is the number of aromatic bonds and $v_{\text{other}}$ is the total bond order from non-aromatic bonds. For example, a carbon atom in benzene typically exhibits configurations like $(2, 1)$—two aromatic bonds and one single bond—or $(3, 0)$, as illustrated in Figure~\ref{fig:aromat_val}. Remarkably, adopting this refined lookup table results in molecular stability scores only 1–3\% lower than originally reported using flawed metrics (third column in Table~\ref{tbl:mol_stab}). While modest, this deviation can meaningfully influence the comparative assessment of generative models and may introduce bias into subsequent benchmark studies if left uncorrected.


An alternative approach involves retraining models on a reprocessed dataset consisting exclusively of kekulized molecules, thereby completely removing ambiguity associated with aromaticity in valency computation. We prepared a revised version of the GEOM-Drugs dataset so that all molecules were kekulized; their is no explicit modeling of aromatic bonds. 
As illustrated in Table~\ref{tbl:mol_stab}, models trained on the kekulized dataset exhibited molecular stability comparable to previously published results when valencies were computed correctly. Notably, all models except Megalodon Flow demonstrated an average 5\% improvement in validity. Megalodon Flow did not show similar improvements. We hypothesize that this discrepancy arises due to smaller neural network architecture used for Megalodon Flow, a decision necessitated by limited computational resources available for this study.



We encountered another issue with GEOM-Drugs: recomputing the valency table on the raw GEOM-Drugs dataset revealed unusual valencies resulting from rare failure in the GFN2-xTB geometry optimization step used to produce the dataset. These failures produced fragmented molecules and unstable valencies such as hydrogen atom with no covalent bonds or neutral carbon with a valency of two.  Examples of these instances are shown in Figure
\ref{fig:broken_mols}. We removed molecules from GEOM-Drugs that were fragmented into multiple disconnected components due to failed GFN2-xTB geometry optimization. This led to the exclusion of 0.18\% of the dataset; although this is not enough data to significantly impact model performance, the presence of these molecules alters the resulting valency lookup table.

To summarize, neither treating aromatic bonds as contributing a valence of 1 nor 1.5 yields chemically accurate results. By correcting the valency table using a refined tuple representation, which captures the number of aromatic bonds separately, the resulting molecular stability scores decrease modestly by 1 to 3\%. However, since most reported stability values exceed 0.9, even such small discrepancies can have an outsized influence, potentially skewing model development and encouraging optimization against a chemically flawed metric. Notably, retraining models on a reprocessed dataset with Kekulized molecules, i.e., without explicit aromatic bonds, leads to approximately a 5\% improvement in validity for 4 of 6 evaluated models. Together, these results underscore the critical importance of chemically sound preprocessing and robust evaluation protocols in the development of 3D molecular generative models.


We make available in the attached github repository the filtered GEOM-Drugs dataset with kekulized molecules, the scripts for producing the filtered dataset from the original GEOM dataset, and an implementation of the molecular stability metric that does not permit erroneous atomic valencies.

\begin{table}[ht]
    \centering
    \begin{threeparttable}
    \footnotesize
        \begin{tabular}{l|cccc|cc}
        \toprule
        \textbf{Model} & \makecell{MS\\ Original} & \makecell{MS\\1.5 Arom} & \makecell{MS\\Arom-Dependent \\Valence} & V\&C & \makecell{MS} & V\&C \\
        \midrule
        EQGAT~\cite{le2023navigating}              & 0.935$^{\pm 0.007}$ & 0.451$^{\pm 0.006}$ & 0.899$^{\pm 0.007}$ & 0.834$^{\pm 0.009}$ & 0.878$^{\pm 0.007}$ & 0.891$^{\pm 0.010}$ \\
        JODO~\cite{huang2024learning}               & 0.981$^{\pm 0.001}$ & 0.517$^{\pm 0.012}$ & 0.963$^{\pm 0.005}$ & 0.879$^{\pm 0.003}$ & \textsuperscript{*}0.940$^{\pm 0.003}$ & \textsuperscript{*}0.923$^{\pm 0.004}$ \\
        Megalodon-quick~\cite{reidenbach2024applications}    & 0.961$^{\pm 0.003}$ & 0.496$^{\pm 0.017}$ & 0.944$^{\pm 0.003}$ & 0.900$^{\pm 0.007}$ & 0.957$^{\pm 0.006}$ & 0.962$^{\pm 0.005}$ \\
        \midrule
        SemlaFlow~\cite{irwinsemlaflow}          & 0.980$^{\pm 0.012}$ & 0.608$^{\pm 0.027}$ & 0.969$^{\pm 0.012}$ & 0.920$^{\pm 0.016}$ & 0.974$^{\pm 0.012}$ & 0.975$^{\pm 0.008}$ \\
        FlowMol2~\cite{dunn2024exploring} & 0.959$^{\pm 0.007}$ & 0.594$^{\pm 0.009}$ & 0.944$^{\pm 0.007}$ & 0.746$^{\pm 0.010}$ & 0.938$^{\pm 0.005}$ & 0.861$^{\pm 0.012}$ \\
        Megalodon-flow~\cite{reidenbach2024applications}     & 0.990$^{\pm 0.003}$ & 0.632$^{\pm 0.011}$ & 0.987$^{\pm 0.004}$ & 0.948$^{\pm 0.003}$ & \textsuperscript{**}0.958$^{\pm 0.004}$ & \textsuperscript{**}0.949$^{\pm 0.002}$ \\
        \bottomrule
        \end{tabular}
\caption{
Comparison of molecular stability (MS) and connected validity (V\&C) across models and processing pipelines. 
The left section reports results obtained using the original GEOM-Drugs dataset and evaluation code: 
``Original'' denotes the values from metric implementations published in prior work, ``1.5 Arom'' reflects scores if aromatic bonds contribute 1.5 to valency, 
and ``Arom-Dependent Valence'' shows scores based on valency computed as $(n_{\text{arom}}, v_{\text{other}})$. 
The right section presents results obtained by retraining on fully Kekulized molecules. V\&C (Valid \& Connected) refers to the fraction of molecules that are both chemically valid and consist of a single connected component.
}
\label{tbl:mol_stab}
\begin{tablenotes}
    \footnotesize
    \item[*] JODO was trained with the EQGAT-Diff objective, using categorical diffusion instead of the original Gaussian formulation for categorical variables.
    \item[**] Indicates results from a retrained “quick” variant, differing from the original paper which reported results for a larger model.
    \end{tablenotes}
    
\end{threeparttable}
\end{table}

\section{3D Molecule Evaluation}

\subsection{Challenges in proper and accurate 3D structure assesment} 

Current 3D molecular generative models face significant challenges in evaluating the geometric quality of their outputs. In particular, models trained on the GEOM-Drugs dataset often exhibit issues stemming from the evaluation protocols themselves. A widely used approach involves defining a bond length lookup table and applying fixed thresholds to assess 3D molecular stability~\cite{huang2024learning, garcia2021n, hoogeboom2022equivariant, morehead2024geometry,song2023equivariant,xu2024geometric}. However, this method proves problematic for GEOM-Drugs: only 86.5\% of atoms satisfy these atom to atom distances, resulting in only 2.8\% of molecules passing the stability criterion.

Our analysis identified just 272 fragmented molecules in the dataset, indicating that geometry optimization with GFN2-xTB converged successfully for the vast majority of conformers. Thus, the observed bond lengths reflect the energy landscape of GFN2-xTB, which may differ from values derived from other sources such as the Cambridge Structural Database (CSD). Despite these discrepancies and the implausibly low stability rates produced by this metric, it remains widely adopted and continues to be propagated in new studies—underscoring the need for a more chemically faithful evaluation standard.

A more recent trend is to assess geometric quality by comparing distributions of bond lengths and angles using Wasserstein distance between generated and training data~\cite{vignac2023midi, le2023navigating, cremer2024pilot}. While this approach is more principled, distributional metrics can be difficult to interpret—particularly outside the computer science community—making it harder to extract chemically meaningful insights.

Other studies have proposed evaluating generated molecules by computing the relaxation energy using molecular mechanics force fields such as MMFF~\cite{xu2022geodiff, irwinsemlaflow,cornet2024equivariant}. However, the choice of force field is critical. For conformers optimized with GFN2-xTB (as in GEOM-Drugs), the mean relaxation energy difference $\Delta E_{\text{relax}}$ when re-optimized with GFN2-xTB is close to zero, as expected. In contrast, the same structures evaluated with MMFF show a mean $\Delta E_{\text{relax}}$ of around $16\,\text{kcal/mol}$, consistent with prior reports of MMFF errors in the $15$–$20\,\text{kcal/mol}$ range relative to higher-level methods\citep{foloppe2019energy}.

As we will demonstrate, current state-of-the-art generative models can now outperform MMFF precision on GEOM-Drugs in terms of alignment with GFN2-xTB. This renders MMFF-based comparisons unreliable and masks meaningful differences between models. However, MMFF energy can still serve as a coarse-grained filter to eliminate structurally implausible molecules, similar to its use in PoseBusters~\cite{buttenschoen2024posebusters} for energy-based outlier detection.
Given the widespread reliance on inadequate metrics, we argue that a GFN2-xTB-based evaluation pipeline is necessary for accurately assessing the practical performance of 3D molecular generative models.

\subsection{GFN2-xTB energy-based geometry benchmark}

Since the geometries of GEOM-Drugs dataset are optimized with GFN2-xTB semi-emperical quantum calculation method, it is essential to use the same energy evaluation method to assess structural integrity of generated molecules. One approach is to measure of how close a generated structure is to the closest local minima of the given energy function. To measure this we suggest to assess differences in bond lengths, bond angles, and torsion angles of generated and optimized counterparts. These quantities provide clear and interpretable measure of generated molecules for both computer scientists and computational chemists.

\paragraph{Bond Length Differences}

For each bond in the molecule, we compute the difference in bond lengths between the initial (generated) and optimized (relaxed) structures. Let $ r_{ij}^{\text{init}}$ and $r_{ij}^{\text{opt}}$ denote the distances between atoms $i$ and $j$ in the initial and optimized conformations, respectively. The bond length difference $\Delta r_{ij}$ is calculated as:

\begin{equation*}
    \Delta r_{ij} = \left| r_{ij}^{\text{init}} - r_{ij}^{\text{opt}} \right|
\end{equation*}

The average difference is reported as a result. 

\paragraph{Bond Angle Differences}

For each bond angle formed by three connected atoms $i$, $j$, and $k$, we calculate the angle difference between the initial and optimized structures. Let $ \theta_{ijk}^{\text{init}}$ and $ \theta_{ijk}^{\text{opt}} $ represent the bond angles at atom $j$ in the initial and optimized conformations, respectively. The bond angle difference $\Delta \theta_{ijk}$ is given by:

\begin{equation*}
\Delta \theta_{ijk} = \min\left( \left| \theta_{ijk}^{\text{init}} - \theta_{ijk}^{\text{opt}} \right|, 180^\circ - \left| \theta_{ijk}^{\text{init}} - \theta_{ijk}^{\text{opt}} \right|\right)
\end{equation*}

As with bond lengths, the average difference is reported as a result. 

\paragraph{Torsion Angle Differences}

Torsion angles involve four connected atoms $i$, $j$,  $k$, and $l$. We compute the difference in torsion angles between the initial and optimized structures using:

\begin{equation*}
\Delta \phi_{ijkl} = \min\left( \left| \phi_{ijkl}^{\text{init}} - \phi_{ijkl}^{\text{opt}} \right|,\ 360^\circ - \left| \phi_{ijkl}^{\text{init}} - \phi_{ijkl}^{\text{opt}} \right| \right)
\end{equation*}

where $\phi_{ijkl}^{\text{init}}$ and $\phi_{ijkl}^{\text{opt}}$ are the dihedral angles in the initial and optimized conformations, respectively. This formula accounts for the periodicity of dihedral angles, ensuring the smallest possible difference is used.

The average difference is reported as a result. 

\paragraph{Results}

We report results for EQGAT, Megalodon-quick, SemlaFlow, FlowMol2, and Megalodon-flow, including both the median and mean relaxation energy $\Delta E_{\text{relax}}$—the energy difference between the initial and GFN2-xTB-optimized structures—as well as structural displacement metrics discussed above (see Table~\ref{tbl:energy_metrics}). For each model, 5,000 molecules were evaluated, and a randomly selected subset of 5,000 molecules from GEOM-Drugs was used for baseline comparisons. To compute confidence intervals, all metrics were calculated across five equal-sized splits of 1,000 molecules each. The Table~\ref{tbl:energy_metrics} row labeled “MMFF $\rightarrow$ GFN2-xTB” quantifies the geometric and energetic discrepancies between MMFF-optimized structures and their GFN2-xTB-optimized counterparts, highlighting the structural divergence between force-field and semi-empirical optimization methods. These results clearly demonstrate that diffusion-based models already surpass MMFF in structural precision. Furthermore, we observe a consistent performance gap between flow-matching and diffusion-based models—even when the underlying architecture remains the same—a discrepancy that has not been previously emphasized in the literature. This finding suggests that earlier conclusions may have been influenced by the limited precision of prior evaluation methodologies.

\begin{table}[!ht]
\caption{Energy relaxation and geometric deviation metrics across generative models. Bond lengths ({\AA}), angles (degrees), and energies (kcal/mol) are reported for valid molecules only. Diffusion-based models use 500 steps; flow-matching models use 100 steps. $\Delta E_\text{relax}$ denotes the energy difference between the initial and GFN2-xTB-optimized structures (i.e., the generative model’s deviation from the reference energy landscape). $\Delta E_\text{relax}^\text{MMFF}$ denotes the MMFF94 energy difference between the initial structure and the structure optimized with MMFF94.
}
\label{tbl:energy_metrics}
\centering
\footnotesize
\begin{tabular}{l l | l l l | l l l}
\toprule
Model &  & \makecell{Bond \\ Length \\ ($\times10^{-2}$)} & \makecell{Bond \\ Angles} & \makecell{Torsions} & \makecell{Median \\ $\Delta E_\text{relax}$} & \makecell{Mean \\ $\Delta E_\text{relax}$} & \makecell{Mean \\ $\Delta E_\text{relax}^\text{MMFF}$} \\
\midrule
\multicolumn{2}{l|}{GEOM-Drugs}         & 0.00$^{\pm 0.001}$ & 0.001$^{\pm 0.001}$ & 0.01$^{\pm 0.01}$ & 0.000$^{\pm 0.0001}$ & 0.001$^{\pm 0.001}$ & 16.4$^{\pm 0.2}$ \\
\multicolumn{2}{l|}{MMFF $\rightarrow$ GFN2-xTB} & 1.12$^{\pm 0.01}$ & 1.22$^{\pm 0.004}$ & 4.89$^{\pm 0.10}$ & 9.84$^{\pm 0.06}$ & 11.4$^{\pm 0.2}$ & 0.00$^{\pm 0.05}$ \\
\midrule
\multicolumn{2}{l|}{EQGAT-diff}              & 1.00$^{\pm 0.04}$ & 1.15$^{\pm 0.03}$ & 8.58$^{\pm 0.11}$ & 6.40$^{\pm 0.20}$ & 11.1$^{\pm 0.8}$ & 28.4$^{\pm 1.2}$ \\
\multicolumn{2}{l|}{JODO}                    & 0.77$^{\pm 0.01}$ & 0.83$^{\pm 0.00}$ & 6.01$^{\pm 0.07}$ & 4.74$^{\pm 0.15}$ & 7.04$^{\pm 0.20}$ & 22.1$^{\pm 0.2}$ \\
\multicolumn{2}{l|}{Megalodon}               & 0.66$^{\pm 0.02}$ & 0.71$^{\pm 0.01}$ & 5.58$^{\pm 0.11}$ & 3.19$^{\pm 0.12}$ & 5.76$^{\pm 0.27}$ & 21.6$^{\pm 0.3}$ \\
\midrule
\multicolumn{2}{l|}{SemlaFlow}               & 3.10$^{\pm 0.23}$ & 2.06$^{\pm 0.17}$ & 6.05$^{\pm 0.56}$ & 32.3$^{\pm 3.3}$ & 91.0$^{\pm 21.7}$ & 69.6$^{\pm 9.2}$ \\
\multicolumn{2}{l|}{FlowMol2}                 & 1.30$^{\pm 0.04}$ & 1.62$^{\pm 0.02}$ & 15.0$^{\pm 0.3}$ & 17.9$^{\pm 0.5}$ & 24.3$^{\pm 0.8}$ & 39.4$^{\pm 1.2}$ \\
\multicolumn{2}{l|}{Megalodon-flow}          & 2.30$^{\pm 0.02}$ & 1.62$^{\pm 0.02}$ & 5.58$^{\pm 0.19}$ & 20.9$^{\pm 0.8}$ & 46.9$^{\pm 8.6}$ & 45.5$^{\pm 2.0}$ \\
\bottomrule
\end{tabular}
\end{table}






\section{Conclusion}

In this study, we revisited the GEOM-Drugs benchmark and uncovered several issues in current 3D molecular generative model evaluation pipelines. We demonstrated that widely adopted stability metrics are affected by code errors, chemically inconsistent valency tables, and reliance on postprocessed molecules, leading to inflated model performance. Furthermore, our findings suggest that MMFF-based energy benchmarks may no longer be appropriate for evaluating models trained on GFN2-xTB-optimized structures, as generative models now appear to surpass MMFF in alignment with the reference energy landscape.

To address these limitations, we proposed a refined evaluation protocol incorporating chemically sound valency definitions and GFN2-xTB-based energy and geometry assessments. Our experiments demonstrate that these corrections impact reported performance while preserving the relative rankings of models. Conversely, a high-quality dataset (error-free structures, consistent features, trustworthy labels) and relevant metrics (e.g. appropriate choice of level of theory or realistic valency lookup table) provide a solid foundation that can markedly improve model performance. We hope that this study will raise awareness about importance of chemical structure curation and processing. We believe these improvements will foster more reliable, interpretable, and chemically meaningful progress in 3D molecular generative modeling. Our recommended evaluation methods and GEOM-Drugs processing scripts are available at \url{https://github.com/isayevlab/geom-drugs-3dgen-evaluation}.

\section{Acknowledgement}

O.I. acknowledges support by the NSF grant CHE-2154447. This work used Expanse at SDSC and Delta at NCSA through allocation CHE200122 from the Advanced Cyberinfrastructure Coordination Ecosystem: Services \& Support (ACCESS) program, which is supported by NSF grants \#2138259, \#2138286, \#2138307, \#2137603, and \#2138296.

I.D. and D.K. acknowledge support through R35GM140753 from the National Institute of General Medical Sciences. The content is solely the responsibility of the authors and does not necessarily represent the official views of the National Institute of General Medical Sciences or the National Institutes of Health.

\input{output.bbl}

\clearpage
\section*{Supplementary Information}

\input{si}

\end{document}

%% file: output.bbl
\providecommand{\latin}[1]{#1}
\makeatletter
\providecommand{\doi}
  {\begingroup\let\do\@makeother\dospecials
  \catcode`\{=1 \catcode`\}=2 \doi@aux}
\providecommand{\doi@aux}[1]{\endgroup\texttt{#1}}
\makeatother
\providecommand*\mcitethebibliography{\thebibliography}
\csname @ifundefined\endcsname{endmcitethebibliography}  {\let\endmcitethebibliography\endthebibliography}{}

%% file: si.tex
\subsection{Appendix I: Valency Lookup Tables for Stability Evaluation}

To support rigorous evaluation of 3D molecular generative models, we include here a collection of empirical valency tables derived from the GEOM-Drugs dataset. These tables are used to define chemically plausible bonding patterns, detect invalid topologies, and serve as standardized references for assessing molecular stability in raw generated molecules.

\paragraph{Table~\ref{tab:accurate_valencies}: Allowed Valencies.} This table summarizes the allowed valencies (i.e., number of bonds including hydrogens) observed in valid GEOM-Drugs structures. It lists configurations by element and formal charge. These values are used as a reference for atom-level and molecule-level stability metrics.

\paragraph{Table~\ref{tab:legacy_bad_valencies}: Legacy and Invalid Valencies.} This table contains valencies found in earlier versions of generative model evaluation pipelines, which include chemically implausible or legacy entries due to preprocessing bugs or failed optimization. It is frequently used to benchmark the quality of generated molecules and identify invalid valency assignments. Many recent studies reference or reuse this table directly.

\paragraph{Table~\ref{tab:geom_drugs_tuple_valencies_hier}: Aromatic Valency Tuples.} This table enumerates all observed combinations of aromatic and non-aromatic bonds per element and charge in the dataset. Each entry is represented as a tuple , where  is the count of aromatic bonds and  is the total bond order from non-aromatic bonds. These tuples capture valency patterns that are otherwise ambiguous under standard counting, especially in polyaromatic and heterocyclic systems.

Together, these tables offer a robust and chemically grounded framework for interpreting stability metrics and ensuring consistency in the evaluation of 3D molecule generation pipelines. Table~\ref{tab:legacy_bad_valencies} in particular is widely used in existing benchmarking literature and reproduced here for completeness.

\begin{table}[H]
\centering
\caption{Valency configurations derived from the GEOM-Drugs dataset, organized by element and formal charge.  
Each cell lists the allowed valencies (including implicit hydrogens) observed for a given formal charge.}
\label{tab:accurate_valencies}
\footnotesize
\begin{tabular}{lcccccc}
\toprule
Element & Charge $-2$ & Charge $-1$ & Charge $0$ & Charge $+1$ & Charge $+2$ & Charge $+3$ \\
\midrule
H   & -- & --       & 1       & --      & --      & -- \\
B   & -- & 4        & 3       & --      & --      & -- \\
C   & -- & 3        & 4       & 3       & --      & -- \\
N   & 1  & 2        & 3       & 4       & --      & -- \\
O   & -- & 1        & 2       & 3       & --      & -- \\
F   & -- & --       & 1       & --      & --      & -- \\
Si  & -- & --       & 4       & 5       & --      & -- \\
P   & -- & --       & 3,\,5   & 4       & --      & -- \\
S   & -- & 1        & 2,\,3,\,6 & 3       & 4       & 2,\,5 \\
Cl  & -- & --       & 1       & 2       & --      & -- \\
Br  & -- & --       & 1       & 2       & --      & -- \\
I   & -- & --       & 1       & 2       & 3       & -- \\
Bi  & -- & --       & 3       & --      & 5       & -- \\
\bottomrule
\end{tabular}
\end{table}

\begin{table}[H]
\centering
\caption{Historically used but chemically implausible valency configurations by formal charge.  
This reference table has been widely used to assess molecular generative models.  
Values highlighted in \textcolor{red}{red} represent known incorrect or unstable configurations; 
values highlighted in \textcolor{blue}{blue} were missing from historical tables but are observed in the dataset.}
\label{tab:legacy_bad_valencies}
\footnotesize
\begin{tabular}{lcccccc}
\toprule
Element & Charge $-2$ & Charge $-1$ & Charge $0$ & Charge $+1$ & Charge $+2$ & Charge $+3$ \\
\midrule
H   & –   & \textcolor{red}{0} & 1             & \textcolor{red}{0}    & –       & –       \\
B   & –   & \textcolor{blue}{4} & 3             & –       & –       & –       \\
C   & –   & 3                  & \textcolor{red}{3},\,4 & 3  & –       & –       \\
N   & 1\textcolor{blue}{} & 2     & \textcolor{red}{2},\,3 & \textcolor{red}{2},\,\textcolor{red}{3},\,4 & – & – \\
O   & –   & 1                  & 2             & 3       & –       & –       \\
F   & –   & \textcolor{red}{0} & 1             & –       & –       & –       \\
Al  & –   & –                  & 3             & –       & –       & –       \\
Si  & –   & –                  & 4             & \textcolor{blue}{5} & –       & –       \\
P   & –   & –                  & 3,\,5         & 4       & –       & –       \\
S   & –   & \textcolor{blue}{1},\,\textcolor{red}{3} & 2,\,6  & \textcolor{red}{2},\,3 & 4    & 5 \\
Cl  & –   & –                  & 1             & 2       & –       & –       \\
Br  & –   & –                  & 1             & 2       & –       & –       \\
Se  & –   & –                  & 2,\,4,\,6     & –       & –       & –       \\
I   & –   & –                  & 1             & 2       & \textcolor{blue}{3} & – \\
Hg  & –   & –                  & 1,\,2         & –       & –       & –       \\
Bi  & –   & –                  & 3             & –       & 5       & – \\
\bottomrule
\end{tabular}
\end{table}

\bigskip

\begin{table}[H]
\centering
\caption{Allowed valency combinations by element and number of aromatic bonds.  
Each cell shows normal valencies for a given atom type and number of aromatic neighbours (row) and formal charge (column). “–” indicates no observed combinations.}
\label{tab:geom_drugs_tuple_valencies_hier}
\footnotesize
\begin{tabular}{l c c c c c c c}
\toprule
Element & \# Aromatic & Charge $-2$ & Charge $-1$ & Charge $0$ & Charge $+1$ & Charge $+2$ & Charge $+3$ \\
\midrule
\multirow{1}{*}{H}
  & 0 & -- & -- & 1 & -- & -- & -- \\
\midrule
\multirow{1}{*}{B}
  & 0 & -- & 4 & 3 & -- & -- & -- \\
\midrule
\multirow{3}{*}{C}
  & 0 & -- & 3 & 4 & 3 & -- & -- \\
  & 2 & -- & 1 & 2,\,1 & 1 & -- & -- \\
  & 3 & -- & 0 & 0 & 0 & -- & -- \\
\midrule
\multirow{4}{*}{N}
  & 0 & 1 & 2 & 3 & 4 & -- & -- \\
  & 2 & -- & 0 & 0,\,1 & 0,\,1,\,2 & -- & -- \\
  & 3 & -- & -- & 0 & 0 & -- & -- \\
\midrule
\multirow{2}{*}{O}
  & 0 & -- & -- & 2 & 3 & -- & -- \\
  & 2 & -- & -- & 0 & -- & -- & -- \\
\midrule
\multirow{1}{*}{F}
  & 0 & -- & -- & 1 & -- & -- & -- \\
\midrule
\multirow{1}{*}{Si}
  & 0 & -- & -- & 4 & 5 & -- & -- \\
\midrule
\multirow{1}{*}{P}
  & 0 & -- & -- & 3,\,5 & 4 & -- & -- \\
\midrule
\multirow{4}{*}{S}
  & 0 & -- & 1 & 2,\,3,\,6 & 3 & 4 & 2,5 \\
  & 2 & -- & -- & 0 & 0,\,1 & -- & -- \\
  & 3 & -- & -- & -- & 0 & -- & -- \\
\midrule
\multirow{1}{*}{Cl}
  & 0 & -- & -- & 1 & 2 & -- & -- \\
\midrule
\multirow{1}{*}{Br}
  & 0 & -- & -- & 1 & 2 & -- & -- \\
\midrule
\multirow{1}{*}{I}
  & 0 & -- & -- & 1 & 2 & 3 & -- \\
\midrule
\multirow{1}{*}{Bi}
  & 0 & -- & -- & 3 & -- & 5 & -- \\
\bottomrule
\end{tabular}
\end{table}

\subsection{Appendix II: Examples of Fractured Compounds in GEOM-Drugs}
\label{sec:broken_mols}

\begin{figure}[H]
  \centering
  \includegraphics[width=0.6\linewidth]{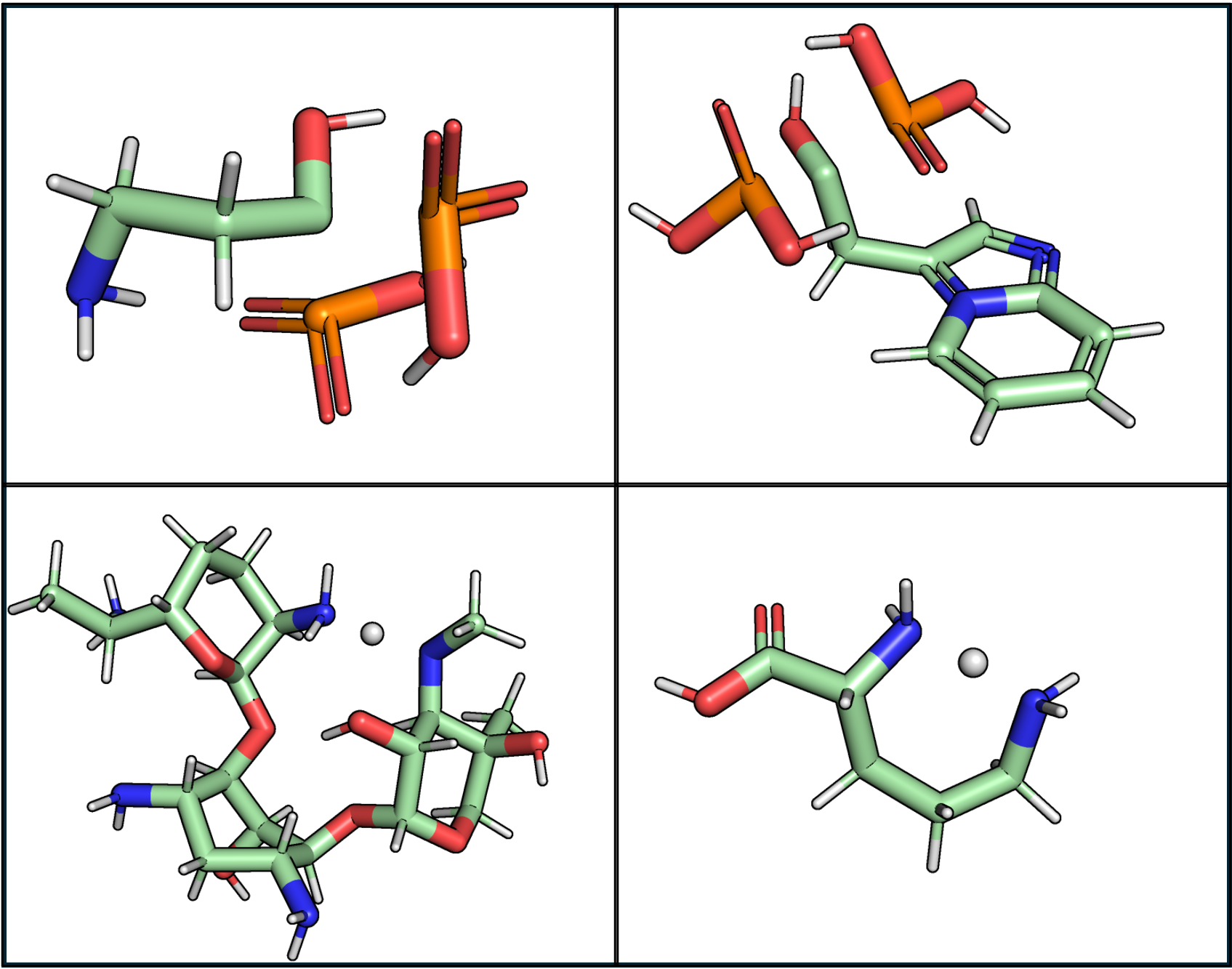}
\caption{
Examples from GEOM-Drugs where GFN2-xTB failed and resulted in fractured molecules. The first row of molecules have neutral carbon with valency 2 and those in the second row have a positively charged hydrogen with valency zero.
}
  \label{fig:broken_mols}
\end{figure}